\documentclass[10pt,twocolumn,letterpaper]{article}

\usepackage{iccv}              
\usepackage{times}
\usepackage{epsfig}
\usepackage{graphicx}
\usepackage{amsmath}

\usepackage{booktabs}
\usepackage{amssymb}
\usepackage{bbding}
\usepackage{pifont}
\usepackage{wasysym}
\usepackage{utfsym}
\usepackage{fontawesome}

\usepackage{makecell}

\usepackage{silence}
\usepackage{multirow}

\newcommand{\namefontcolor}[1]{\textcolor{black}{#1}}
\newcommand{\figref}[1]{Fig.~\ref{#1}}
\newcommand{\tabref}[1]{Tab.~\ref{#1}}
\newcommand{\eqnref}[1]{Eqn.~\ref{#1}}
\newcommand{\secref}[1]{Sec.~\ref{#1}}
\newcommand{\myPara}[1]{\noindent\textbf{#1}}
\newcommand{\sArt}{state-of-the-art~}

\def\eg{\emph{e.g.,~}}
\def\ie{\emph{i.e.,~}}

\newcommand{\nMethod}{AR-1-to-3}
\newcommand{\nameoflocalmodule}{Stacked-LE}
\newcommand{\nameofglobalmodule}{LSTM-GE}

\definecolor{iccvblue}{rgb}{0.21,0.49,0.74}
\usepackage[pagebackref,breaklinks,colorlinks,allcolors=iccvblue]{hyperref}

\def\paperID{1122} 
\def\confName{ICCV}
\def\confYear{2025}

\title{\nMethod{}: Single Image to Consistent 3D Object via Next-View Prediction}

\author{\href{https://zhangxuying1004.github.io/}{\namefontcolor{Xuying Zhang}}$^{1}$\textsuperscript{*}
    \quad \href{https://scholar.google.com/citations?hl=zh-CN\&user=zQTdfUIAAAAJ}{\namefontcolor{Yupeng Zhou}}$^{1}$\textsuperscript{*}
    \quad \href{https://scholar.google.com/citations?user=I6kIOM0AAAAJ\&hl=zh-CN}{\namefontcolor{Kai Wang}}$^{1}$
    \quad \href{https://yikaiw.github.io/}{\namefontcolor{Yikai Wang}}$^{3}$  
    \quad \href{https://paper99.github.io/}{\namefontcolor{Zhen Li}}$^{1}$ \\
    \quad \href{https://scholar.google.com/citations?hl=zh-CN\&user=TiTxZloAAAAJ}{\namefontcolor{Shaohui Jiao}}$^{4}$ 
    \quad \href{https://zhoudaquan.github.io/homepage.io/index.html}{\namefontcolor{Daquan Zhou}}$^{4}$
    \quad \href{https://houqb.github.io/}{\namefontcolor{Qibin Hou}}$^{1,2}$\textsuperscript{\dag} 
    \quad \href{https://mmcheng.net/cmm/}{\namefontcolor{Ming-Ming Cheng}}$^{1,2}$   \\ 
    \small $^1$ VCIP, CS, Nankai University  \quad $^2$NKIARI, Shenzhen Futian \quad $^3$Tsinghua University  \quad $^4$ByteDance Inc. \\
    {\tt\small Homepage:\url{https://github.com/HVision-NKU/AR123}}
}

\begin{document}

\twocolumn[{%
    \renewcommand\twocolumn[1][]{#1}%
    \maketitle
    \vspace{-39pt}
    \begin{center}
      \captionsetup{type=figure}   
      \footnotesize
       \includegraphics[width=0.95\linewidth]{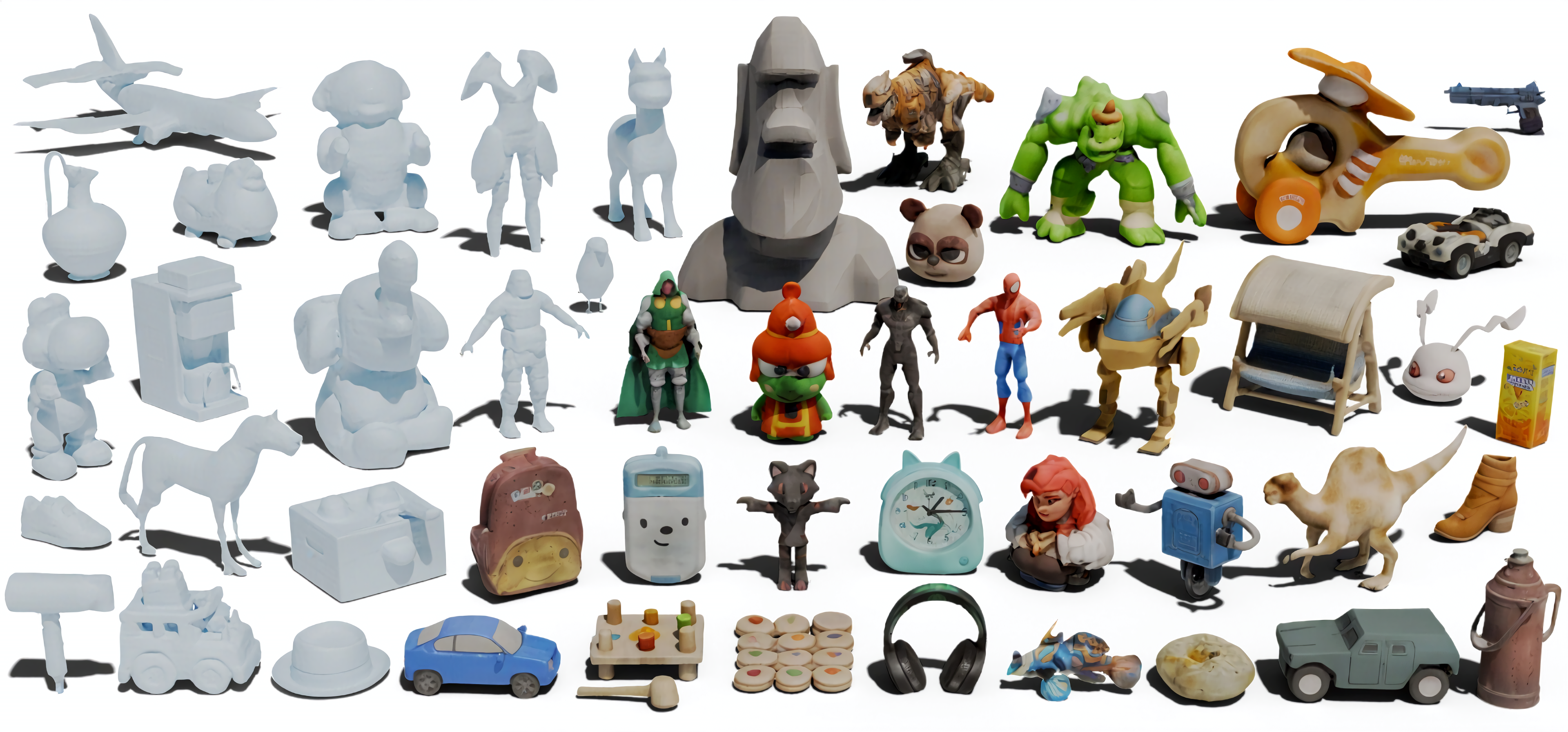}
       \vspace{-13pt}    
      \captionof{figure}{
        Examples of high-quality 3D asset gallery generated by our \nMethod{} model.
        See \secref{subsec:quali} for more visualization results.
      }
      \label{fig:gallery}
    \end{center}%
}]  

\footnotetext[0]{
    \textsuperscript{*}Equal contribution. ~ 
    \textsuperscript{\dag}Corresponding author.
}


\begin{abstract}
    Novel view synthesis (NVS) is a cornerstone for image-to-3d creation.
    However, existing works still struggle to maintain consistency between the generated views and the input views, especially when there is a significant camera pose difference, leading to poor-quality 3D geometries and textures.
    We attribute this issue to their treatment of all target views with equal priority according to our empirical observation that the target views closer to the input views exhibit higher fidelity.
    With this inspiration, we propose \nMethod{}, a novel next-view prediction paradigm based on diffusion models that first generates views close to the input views, which are then utilized as contextual information to progressively synthesize farther views.
    %
    To encode the generated view subsequences as local and global conditions for the next-view prediction, we accordingly develop a stacked local feature encoding strategy (\nameoflocalmodule{}) and an LSTM-based global feature encoding strategy (\nameofglobalmodule{}).
    Extensive experiments demonstrate that our method significantly improves the consistency between the generated views and the input views, producing high-fidelity 3D assets.

\end{abstract}

\section{Introduction}\label{sec:intro}
\vspace{5.5pt}
Synthesizing 3D objects from a single-view image has been a fundamental and enduring research in both computer vision and graphics communities.
This task is challenging, as it not only involves reconstructing the visible regions of a 3D object but also necessitates extrapolating the invisible parts.
To address these challenges, researchers have sought to synthesize novel views~\cite{qian2023magic123,chen2023cascade,long2024wonder3d,liu2024one} with the powerful priors of diffusion models, which have achieved remarkable success in image generation~\cite{rombach2022high,peebles2023scalable,li2024photomaker,zhou2024storydiffusion,chen2024region}.

\begin{figure}[t]
    \centering
    \footnotesize
    \includegraphics[width=\linewidth]{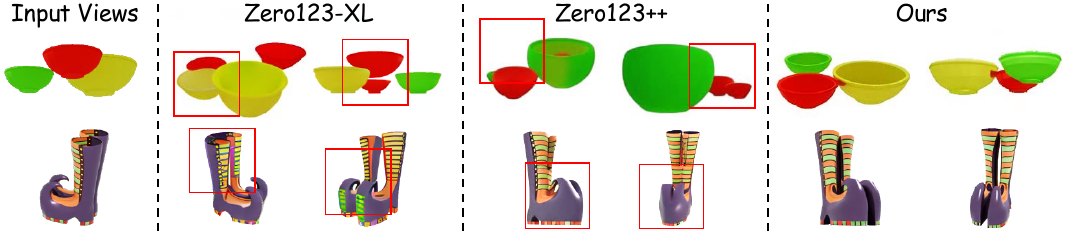}
    \vspace{-19pt}
    \caption{
    Novel view examples generated by different paradigms when encountering significant differences in camera poses.}
    \label{fig:intro}
    \vspace{-15pt}
\end{figure}

Existing approaches can be broadly categorized into two distinct paradigms.
To be specific, the methods represented by Zero123~\cite{liu2023zero} fine-tune pre-trained diffusion models on $<$\emph{input\_view, camera\_pose, output\_view}$>$ triplets of 3D renders and generate multiple discrete views of a 3D object based on a single-view image and a set of camera poses.
On another line, Zero123++~\cite{shi2023zero123++} and its follow-ups~\cite{li2023instant3d,tang2024lgm,xu2024instantmesh,liu2024oneplus} consider the independence of new views specified by different camera poses.
They utilize diffusion models to generate multiple views simultaneously by merging all target views with specified camera poses into a grid image for joint distribution modeling.
Despite their success in many scenarios, these methods remain prone to synthesizing new views inconsistent with the input image, especially when there is a large gap between their camera poses, 
as illustrated in the bowls and shoes of \figref{fig:intro}.

\begin{figure}[t]
    \centering
    \footnotesize
    \includegraphics[width=\linewidth]{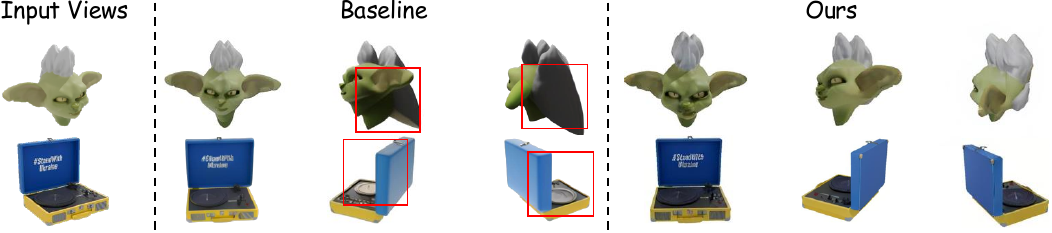}
    \vspace{-20pt}
    \caption{
    Novel view examples of nearby and distant camera poses generated by the Zero123++ baseline and our \nMethod{}.}
    \label{fig:intro1}
    \vspace{-13pt}
\end{figure}

A fundamental reason for this issue is that these methods generate novel views for all camera poses in an equally prioritized manner.
We argue that novel views with camera poses closer to the input view should be generated first in that they usually exhibit substantially higher generation fidelity, whereas those with larger pose variations present greater challenges, as shown in \figref{fig:intro1}.
To this end, we rethink the way to generate consistent views and present \nMethod{}, a novel paradigm to progressively generate all target views, with the closer views generated first serving as contextual information to generate the farther view.

Our methodology follows the 3$\times$2 grid image generation strategy proposed by Zero123++~\cite{shi2023zero123++}.
%
It is noteworthy that there exists a potential sequential relationship between these six target views, where the adjacent rows of views in the grid share identical elevation angles and a fixed azimuth interval of $120^{\circ}$.
This sequential nature enables our \nMethod{} to start from generating the first row views based on the input view and gradually generate the remaining row views in an autoregressive fashion.
%
At each iteration step, the target views from two different camera elevations can exchange information, and the views generated in the previous steps are utilized as references to generate current views.

To encode partially generated sequence views and provide references for the next views, we develop two image conditioning strategies, \ie \nameoflocalmodule{} and \nameofglobalmodule{}, for local conditioning and global conditioning of the diffusion models respectively.
In the \nameoflocalmodule{} strategy, the denoising UNet model encodes the previously generated views into a stack embedding, which is employed as pixel-wise spatial guidance to modify the key and value matrices of the self-attention layers for denoising the target views of the current step.
In \nameofglobalmodule{}, the view subsequence is divided into 
two groups according to the elevation, whose feature vectors are encoded by two LSTM modules for high-level semantic conditions of current views.

We evaluate the performance of our \nMethod{} paradigm on a 3D benchmark dataset, \eg Objaverse~\cite{deitke2023objaverse}, and two out-of-domain datasets, \ie Google Scanned Objects~\cite{downs2022google}, and OmniObject3D~\cite{wu2023omniobject3d}.
By introducing the autoregressive manner coupled with the proposed \nameoflocalmodule{} and \nameofglobalmodule{} strategies into multi-view diffusion models, our \nMethod{} enables consistent and accurate novel view synthesis, resulting in high-quality 3D assets, as shown in \figref{fig:gallery}.
The experimental results also demonstrate the superiority of our \nMethod{} compared to cutting-edge new view synthesis methods and image-to-3d methods.

Our contributions can be summarized as follows:
\begin{itemize}
    \item We propose \nMethod{}, an autoregressive next-view prediction framework for 3D object generation, which can generate target views from near to far progressively.
    \item We design \nameoflocalmodule{} and \nameofglobalmodule{} strategies to encode the partially generated sequence views and provide local and global conditions for the diffusion models.
    \item Extensive quantitative and qualitative experiments on large-scale 3D datasets demonstrate that our approach can generate more consistent 2D multi-view images than previous works and produce high-quality 3D assets.
\end{itemize}

\section{Related Work}
\subsection{2D Diffusion Models for 3D Generation}
Diffusion Models~\cite{rombach2022high,peebles2023scalable,black2024flux,chen2024adaptive,du2025textcrafter} pre-trained on large-scale 2D datasets have demonstrated remarkable performance in generating high-quality images and powerful zero-shot generalization.
In recent years, significant effort has been consecutively devoted to transferring such merits of 2D diffusion models to 3D generation.

Early methods~\cite{poole2022dreamfusion,chen2023fantasia3d,zhang2024temo,li2024director3d,qu2025drag} distill knowledge of 2D pre-trained models by feeding the rendered views to it and performing per-shape optimization but suffer from artifacts such as over-saturated colors and the ``multi-face'' problem.
Zero123~\cite{liu2023zero} pioneers an open-world single-image-to-3D framework, in which diffusion models are fine-tuned to synthesize new views conditioned on an input view and a set of discrete camera poses.
ImageDream~\cite{wang2023imagedream} adopts world camera coordination as in MVDream~\cite{shi2023mvdream} to recover 3D geometries.
Magic123~\cite{qian2023magic123} combines the 3D prior of Zero123 and the 2D prior of the stable diffusion model together to enhance the quality of generated 3D meshes.
One-2-3-45~\cite{liu2024one} uses Zero123 to generate multi-view images, which are lifted to 3D space to assist in the generation of 3D meshes.
Several approaches like Consistent123~\cite{lin2023consistent123} and MVD-Fusion~\cite{hu2024mvd}, improve Zero123 by incorporating additional priors, \ie boundary and depth.
Besides, more and more attention has been drawn to enforce consistency between the generated multiple views.
SyncDreamer~\cite{liu2023syncdreamer} adopts a 3D-aware attention mechanism to correlate the corresponding features across different views.
MVDiffusion~\cite{Tang2023mvdiffusion} generates multi-view images in parallel through the weight-sharing multi-branch UNet with shared weights and correspondence-aware attention. 
Recent methods~\cite{tang2024cycle3d,zheng2024free3d,li2024dual3d,gao2024cat3d} explore to  simultaneously generate multiple views to model their joint distribution.
Zero123++~\cite{shi2023zero123++} further proposes a strategy of tiling six target views surrounding the 3D object into a grid image.
This strategy has been successively adopted by follow-up works such as One-2-3-45++~\cite{liu2024oneplus}, Instant3D~\cite{li2023instant3d}, and InstantMesh~\cite{xu2024instantmesh}.

In contrast to these methods heavily relying on the 2D diffusion priors, our \nMethod{} inspired by human thinking also pays attention to the contextual information of the current object.
The method most similar to ours is Cascade-Zero123~\cite{chen2023cascade}.
It first uses a multi-view diffusion model to generate many extra views, which, along with the input image, are then fed into another diffusion model to produce a specific target view.
Different from this method, our \nMethod{} takes the relationship between the target views and the input image into account and utilizes a diffusion model to establish the potential sequence between them.

\subsection{Autoregressive Generation}

The autoregressive scheme, which operates on the fundamental principle that the current value in a sequence can be explained by its preceding values, plays a vital role in the deep learning era~\cite{mu2023neural,wang2023nice,zhang24FastPCI,wu2025ret3d,lu2025learning}.
Based upon this technology, researchers have developed a series of classical sequential modeling methods~\cite{hochreiter1997long,shi2015convolutional,vaswani2017attention,ma2023towards,li2024atprompt}.
In recent years, there has been a growing interest in extending such an autoregressive fashion to various communities.

PixelRNN~\cite{van2016pixel} is one of the pioneering methods to generate high-quality images with intricate details by modeling pixel dependencies.
VQ-VAE~\cite{van2017neural} revolutionizes the learning of discrete representations by incorporating the codebook mechanism, enabling efficient encoding and decoding processes of images.
VQ-GAN~\cite{esser2021taming} adopts a transformer architecture to model serialized visual parts and introduces adversarial loss during the training process.
Parti~\cite{yu2022scaling} proposes a pathways autoregressive model treating the image generation as a sequence-to-sequence modeling to generate high-fidelity photorealistic images.
VAR~\cite{tian2024visual}, LlamaGen~\cite{sun2024autoregressive}, and Infinity~\cite{han2024infinity} scale image generation by incorporating multimodal large models.
In addition, some approaches~\cite{ruhe2024rolling,jin2024pyramidal,chen2025diffusion,voleti2024sv3d,zuo2024videomv} attempt to integrate diffusion models with the sequential strategy to achieve more temporally consistent video generation.
More recently, MeshGPT series~\cite{siddiqui2024meshgpt,chen2024meshanything,chen2024meshxl} propose to generate triangle faces in an autoregressive manner for artist-created 3D meshes.
TAR3D~\cite{zhang2024tar3d} and SAR3D~\cite{chen2024sar3d} quantify 3D latent representations~\cite{ji2024jm3d}, which are used to generate SDF-based 3D objects~\cite{zhu_2024_gsror,zhu_2025_dsdf} with the next-token prediction strategy.

In this work, we observe the target views of Zero123++ can be divided into several steps with equidistant camera pose intervals, resulting in a potential sequential nature.
%

\section{Methodology}\label{sec:method}

\subsection{Preliminaries}
We introduce the preliminaries of Zero123++~\cite{shi2023zero123++}, the base multi-view diffusion model adopted in this work, which is beneficial for understanding the designs in \nMethod{}.

\myPara{Multi-View Generation.}
To achieve the modeling of the joint distribution between multiple new views, Zero123++ proposes to tile six target views with a $3 \times 2$ layout in a grid image.
Note that these target views are obtained from a fixed set of relative azimuth and absolute elevation angles.
Specifically, they consist of interleaving elevation angles of $20^{\circ}$ downwards and $-10^{\circ}$ upwards, combined with azimuth angles starting from $30^{\circ}$ relative to the input azimuth and incrementing by $60^{\circ}$ for each subsequent camera pose.

\myPara{Stable Diffusion.}
Zero123++ chose Stable Diffusion (SD) as the generative model since it is open-sourced and has been trained on various internet-scale image datasets.
The geometric priors that SD learns about natural images are utilized for novel view synthesis under the image and camera conditions.
SD performs the diffusion process within the latent space of a pre-trained autoencoder whose encoder and decoder are denoted as $\mathcal{E}(\cdot)$and $\mathcal{D}(\cdot)$, respectively.
At the diffusion time step $t$, the objective for fine-tuning the denoiser UNet $\epsilon_{\theta}(\cdot)$ can be formulated as:
\begin{equation} 
   \mathcal{L} = \mathbb{E}_{z \sim \mathcal{E}(x),y,t,\epsilon \sim \mathcal{N}(0,I)}||\\ \epsilon - \epsilon_{\theta}(z_{t},t,c_{\theta}(y) ||\\_{2}^{2},
  \label{eq:zero123plusobj}
\end{equation}
where $x$ is the target grid image which is perturbed to a feature by Gaussian noise $\epsilon$, \ie $z_{t}$, and $c_{\theta}(y)$ represents the embedding encoded from the condition image $y$.

\myPara{Image Condition.}
The image conditioning techniques in Zero123++, \ie $c_{\theta}(y)$, can be divided into local and global conditions. 
The former strategy is tailored for the pixel-wise spatial correspondence between the input and target views.
Specifically, it adopts a variant of the Reference Attention operation~\cite{zhang2023reference}, which runs the denoising UNet on the input image and appends the self-attention key and value matrices from it to the corresponding attention layers when denoising the target views.
Note that the Gaussian noise at the same level as the denoising input is added to the referring image so that the UNet can focus on the relevant features for denoising at the current noise level.
%
In terms of the global conditioning strategy, the CLIP~\cite{radford2021learning} text embedding of an empty text is added with the CLIP image embedding of the input image multiplied by a trainable set of global weights to provide high-level semantic information for the cross-attention of the denoising UNet.

\begin{figure*}[t]
    \centering
    \footnotesize
    \includegraphics[width=0.98\linewidth]{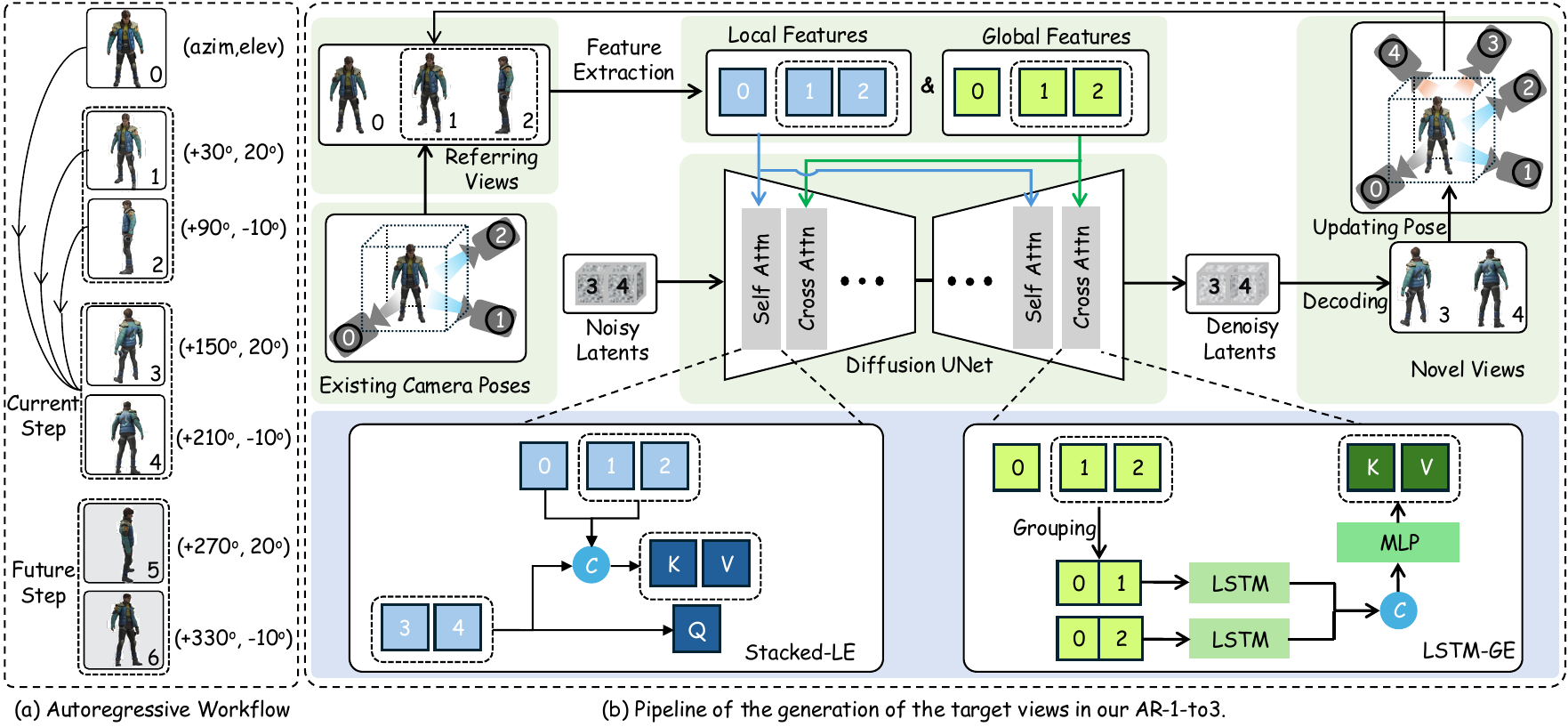}
    \vspace{-1pt}
    \caption{Overview of our \nMethod{} framework.
    The left side shows the \nMethod{} workflow, while the right side illustrates the denoising process of target views.
    Taking the input single-view image as initialization, our methodology employs a diffusion model to generate all target views incrementally from near to far, with the existing views from previous steps serving as contextual information about the objects themselves.
    To achieve this, the \nameoflocalmodule{} and \nameofglobalmodule{} strategies are developed to encode the local features and global features of the partial view sequence as the image conditions of the denoising UNet for the view prediction of the current step.
    }
    \vspace{-3pt}
    \label{fig:overall}
\end{figure*}

\subsection{\nMethod}
Existing methods either generate multiple discrete viewpoints from a single input view and a set of camera poses, like Zero123~\cite{liu2023zero}, One-2-3-45\cite{liu2024one}, or simultaneously generate multiple views in a grid layout based on specified camera conditions, such as Zero123++~\cite{shi2023zero123++}, ImageDream~\cite{wang2023imagedream}.
Despite achieving excellent performance in many scenarios, these methods are still prone to generating several target views that exhibit geometric and textural inconsistencies with the input image.
We argue that their equal prioritization of all new views and the underutilization of contextual information of current objects during the generation process should be responsible for this issue.

The core of this work is to progressively generate target views in a next-view prediction fashion so that the closer views generated earlier can be used as supplementary information for the generation of the farther views.
\figref{fig:overall} shows the end-to-end architecture of our \nMethod{}.
In this work, we follow the paradigm of Zero123++ generating 6 specific camera conditional target views.
In contrast, our method generates these views step by step rather than all at once.
In addition, Zero123++ has demonstrated that generating multiple target views simultaneously contributes to accurately modeling the joint distribution of these views.
Therefore, each step in our generative strategy refers to a row of the $3 \times 2$ layout, containing two target views with different elevations and $60^{\circ}$ difference in azimuth.
Moreover, the difference in camera pose between adjacent steps is a fixed azimuth angle of $120^{\circ}$, which is suitable for the next view prediction scheme.
As a result, the different target views at each step can exchange information, and the target views generated in the previous steps can be utilized as extra conditions to generate the views for the current step.

We achieve such next-view prediction by designing two image conditional strategies that encode sequence view information to fine-tune the diffusion model.
These two strategies, denoted as Stacked Local Feature Encoding (\nameoflocalmodule{}) and Long Short-Term Global Feature Encoding (\nameofglobalmodule{}), correspond to the local and global image conditioning techniques in Zero123++, respectively.
The optimization objective can still be represented by \eqnref{eq:zero123plusobj}, and we will elaborate on our image conditional policy, \ie $c_{\theta}(y)$, in the subsequent \secref{subsec:local} and \secref{subsec:global}.

%
Through multi-step autoregression, our \nMethod{} gradually generates 6 target views, which are fed to a sparse-view large reconstruction model to obtain a 3D object.
In this work, we choose the pre-trained InstantMesh~\cite{xu2024instantmesh} as our 3D reconstruction model, which encodes the multi-view images as triplane features via a transformer-based architecture~\cite{hong2023lrm} and predicts the point color and density for volumetric rendering by a multi-layer perceptron.

\subsection{Stacked Local Feature Encoding} \label{subsec:local}
In this subsection, we introduce how our method encodes the latent features of the input image and the generated partial target-view sequence as local conditions for the operation of Reference Attention to generate the target views of the current step. 
Note that the denoising UNet model is a multi-level architecture and the hidden dimensions may vary across different self-attention layers.
It is challenging to encode the latent features of the condition view sequence at these positions using a single network.
Considering that these reference features and the attention representations in self-attention layers come from the same positions in the U-Net network, sharing the same spatial and channel dimensions, we naturally thought of encoding them into a unified representation by stacking them along the spatial dimension.
This strategy offers two significant advantages: 
1) it can encode any number of reference features at the self-attention layer.
2) it allows the reference features to be directly fed into the attention module, enabling the reuse of weight parameters without any additional design required.
We term this local feature encoding strategy as \nameoflocalmodule{} whose details are shown in the bottom left corner of \figref{fig:overall}.

Formally, at the $k$-th step of autoregression, with a total of $2k$-$1$ reference views, we aim to predict the $(2k)$-th and $(2k$+$1)$-th target views, where the variable $k$ ranges from 1 to 3.
Following Zero123++~\cite{shi2023zero123++}, we first feed the referring views into the denoising UNet model separately and record their key/value matrices at the self-attention module.
%
%
Then, we perform another forward pass with the U-Net to denoise the target views of the current step.
During this process, the records in each layer are stacked together to modify the key and value matrices in the self-attention module of the corresponding layer, which can be defined as:
\begin{equation} 
   s^{*}_{i} = \mathrm{Concat}([e^{1}_{i}, e^{2}_{i}, ..., e^{2k-1}_{i}, s_{i}]),
  \label{eq:local_att_kv}
\end{equation}
where $s_{i} \in \mathbb{R}^{B \times L \times D_{i}}$ denotes the key or value matrices of the $i$-th self-attention, and $e_{i}^{j}$ is the recorded embeddings of the $j$-th reference view.
Note that $B$, $L$ and $D_{i}$ are the batch size, token number, and dimension of the key /value matrices.
Further, we compute the self-attention as follows:
\begin{equation} 
   O_{i} = \mathrm{Attention}([Q_{i}, K^{*}_{i}, V^{*}_{i}]).
  \label{eq:local_att}
\end{equation}


\subsection{Long Short-Term Global Feature Encoding} \label{subsec:global}
In this subsection, we present the details of encoding the CLIP features of the input images and the existing target-view sequence as global conditions, which provide high-level semantic information via cross-attention to generate the target views of the current step.
We empirically observe that the CLIP features of the conditional view sequence are vectors with the same channel dimension and 1D spatial dimensions.
In addition, they can be divided into two sub-sequences with an azimuthal angle spacing of $120^{\circ}$ based on their elevation angles.
Thus, it is well-suited to process such sequences with the Long Short-Term Memory (LSTM) Network~\cite{hochreiter1997long}.
Furthermore, this manner has two interesting merits, mitigating computational burden: 
1) It can encode feature vectors from the conditional views into two vectors,  regardless of the number of views.
2) The LSTM structure possesses a strong ability to model sequential representations while requiring fewer parameters.
This strategy is named as \nameofglobalmodule{}, and the diagrammatic details are shown in the bottom left corner of \figref{fig:overall}.

Given $2k$-$1$ conditional views at the $k$-th step of autoregression, we first send them to the image encoder of the CLIP model for their visual features, which are represented as $F \in \mathbb{R}^{B \times (2k-1) \times D}$.
Then, we partition these features into two groups according to the elevation angles of their respective views.
Note that the feature of the input image is a special case, which is included in both groups.
We denote the two grouped features as $F_{1} \in \mathbb{R}^{B \times k \times D}$ and $F_{2} \in \mathbb{R}^{B \times k \times D}$, and feed them to two separate LSTM modules.
Next, the hidden states of the $k$-th step of the two LSTM modules, \ie $h_{l}^{k}$, are selected as their respective outputs, \ie, $I_{l}$.
This process can be formulated as follows:
\begin{equation} 
  I_{l} = \mathrm{LSTM}_{l}([F_{l}, (h_{l}^{0}, c_{l}^{0})]), \quad l \in \{0, 1\}
 \label{eq:global_lstm}
\end{equation}
where $h_{l}^{0}$ and $c_{l}^{0}$ are the hidden state and cell state, respectively, both initialized as zero vectors.
Finally, the outputs of the two LSTM modules are concatenated together along the channel dimension, followed by an MLP layer and a trainable set of global weights $W \in \mathbb{R}^{77 \times 1}$, to form global embeddings for the cross-attention of the denoising UNet:
\begin{equation} 
  T = W \cdot \mathrm{MLP}(\mathrm{Concat}([I_{0}, I_{1}])). \quad T \in \mathbb{R}^{B \times 77 \times D}
 \label{eq:global_concat}
\end{equation}
Note that we remove the CLIP embedding of empty text from the original global condition, as empirical observations indicate its negligible impact on the final results.

\section{Experiments}\label{sec:experiment}

\begin{figure*}[t]
    \centering
    \footnotesize
    \includegraphics[width=0.98\linewidth]{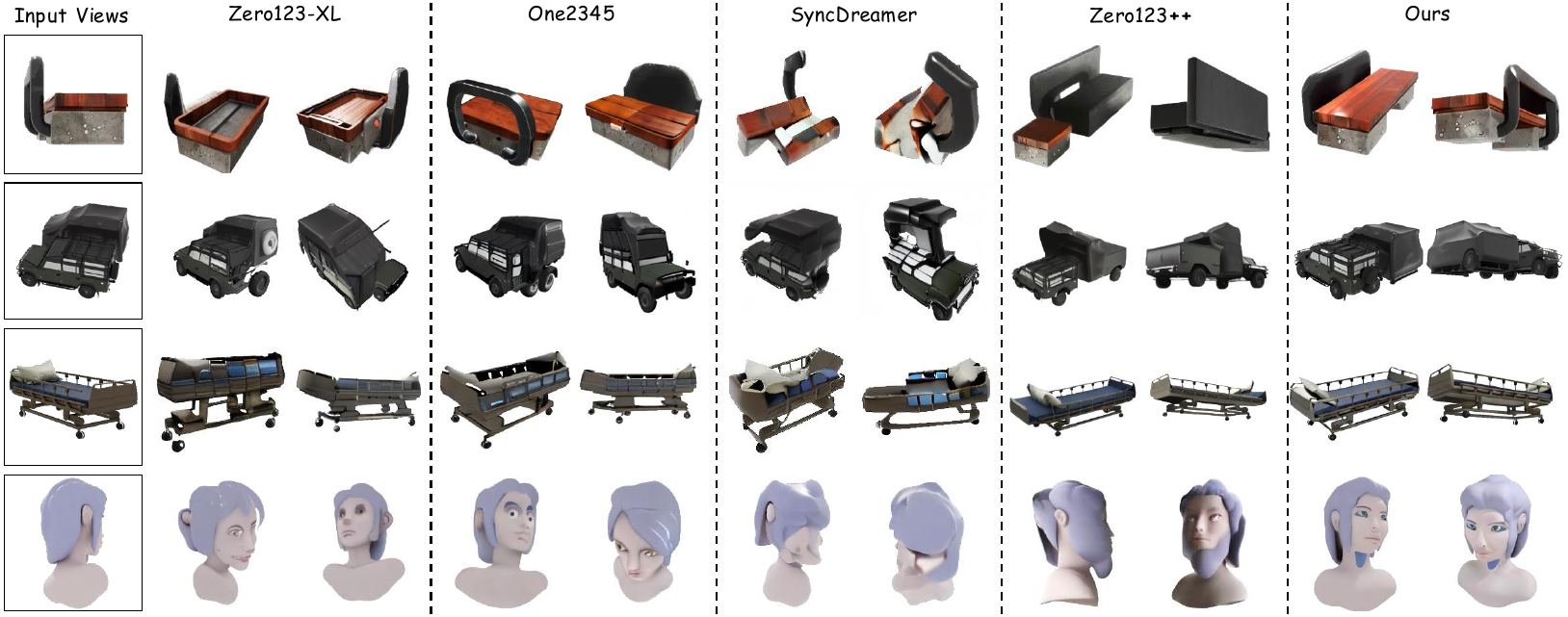}
    \vspace{-7pt}
    \caption{
    Visual comparisons of the novel views synthesized by our \nMethod{} and recent popular methods on multi-view generation.
    Compared with the existing approaches, the new views from our \nMethod{} are more consistent with both each other and the input views.}
    \label{fig:view_syn}
    \vspace{-10pt}
\end{figure*}

    

\begin{figure}[t]
    \vspace{5pt}
    \centering
    \footnotesize
    \includegraphics[width=0.98\linewidth]{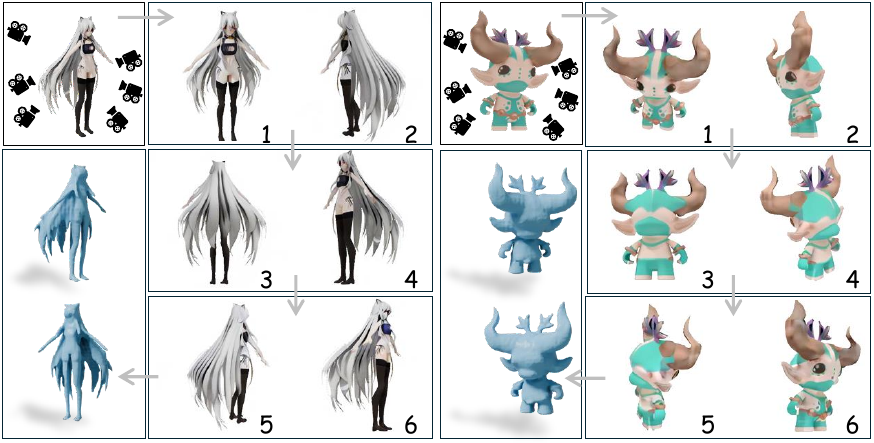}
    \vspace{-7pt}
    \caption{
    Examples of image-to-3d generation based on our next-view prediction.
    \nMethod{} produces multi-view images consistent with the input images, leading to high-quality 3D results.
    }
    \label{fig:show}
    \vspace{-13pt}
\end{figure}
\subsection{Experiment Setup}\label{sec:exp_setting}

\myPara{Datasets.}
We conduct experiments on the most popular benchmarking 3D dataset, \ie Objaverse~\cite{deitke2023objaverse}, along with two out-of-domain datasets, \ie  Google Scanned Objects (GSO)~\cite{downs2022google} and OmniObject3D (Omni3D)~\cite{wu2023omniobject3d}.
Following the filtering principle in previous works~\cite{zhang2024tar3d,li2025triposg}, we obtain about 210,000 geometry objects from the Objaverse dataset with a 3D mesh capacity of 800,000. 
We pick out 300 objects covering various categories to evaluate the performance of the methods, and the remaining samples are used for training.
Moreover, we also randomly select 300 samples each from both GSO and Omni3D to ensure a fair evaluation of the methods involved in this paper.

Following the protocol of Zero123++~\cite{shi2023zero123++}, we render 7 images for each object, including an input image and 6 target images.
To be specific, an elevation angle ranging from $-20^{\circ}$ to $45^{\circ}$ and an azimuth angle ranging from $0^{\circ}$ to $360^{\circ}$ are randomly sampled to render the input image.
The camera poses of the 6 target images involve interleaving absolute elevations of $20^{\circ}$ and $-10^{\circ}$, paired with azimuths relative to the input image that start at $30^{\circ}$ and increase by $60^{\circ}$ for each pose.
Besides, all rendered images are set to a white background to ensure that the diffusion model produces images of this nature, thereby avoiding the trouble of removing the background when reconstructing 3D objects.
We will open-source these rendered images in our project.

\myPara{Evaluation.}
We evaluate the performance of the methods in two critical dimensions, \ie 2D visual fidelity and 3D geometric accuracy.
Specifically, we compare the novel views generated by multi-view diffusion models or rendered from synthesized 3D meshes with the ground truth views.
Following image comparison~\cite{lin2023consistent123,chen2023cascade,xu2024instantmesh}, we report four popular metrics, including Peak Signal-to-Noise Ratio (PSNR), Perceptual Loss(LPIPS)~\cite{zhang2018unreasonable}, Structural Similarity (SSIM)~\cite{wang2004image}, and CLIP-score~\cite{radford2021learning}.
We also compare the surface points randomly sampled from the
generated 3D meshes with those from ground-truth meshes, in which the chamfer distance (CD) value and the F-Score with a threshold of 0.02 are employed as metrics.

\myPara{Implementation Details.}
We train \nMethod{} on the render images from about 210K objects of the Objaverse dataset for 150k steps with a total batch size of 32 on 8 NVIDIA A100 (80G) GPUs.
The learning rate is initialized as $1e$-$5$ and changes every 25k steps in a cycle, along with the AdamW optimizer~\cite{loshchilov2019decoupled} and CosineAnnealingWarmRestarts scheduler~\cite{loshchilov2016sgdr}.
We randomly select a $k$ from $\{1, 2, 3 \}$ to build the autoregressive pattern, where the first $2k$-$1$ views form the conditional images and the following two views make up the target views.
We resize the size of the conditional images to a value between 128 and 512 so that the model is capable of adjusting to different input resolutions and producing more clear images.
Meanwhile, we resize each target view to 320, thereby the size of the grid image during the autoregressive process is 320$\times$640. 
In addition, we employ the linear noise schedule and v-prediction loss in Zero123++~\cite{shi2023zero123++} rather than the alternatives in the Stable Diffusion model~\cite{rombach2022high}.
During the inference stage, taking the input image as initialization, our \nMethod{} generates all target views in three steps, as shown in \figref{fig:show}.

\begin{figure*}[t]
    \centering
    \footnotesize
    \includegraphics[width=0.98\linewidth]{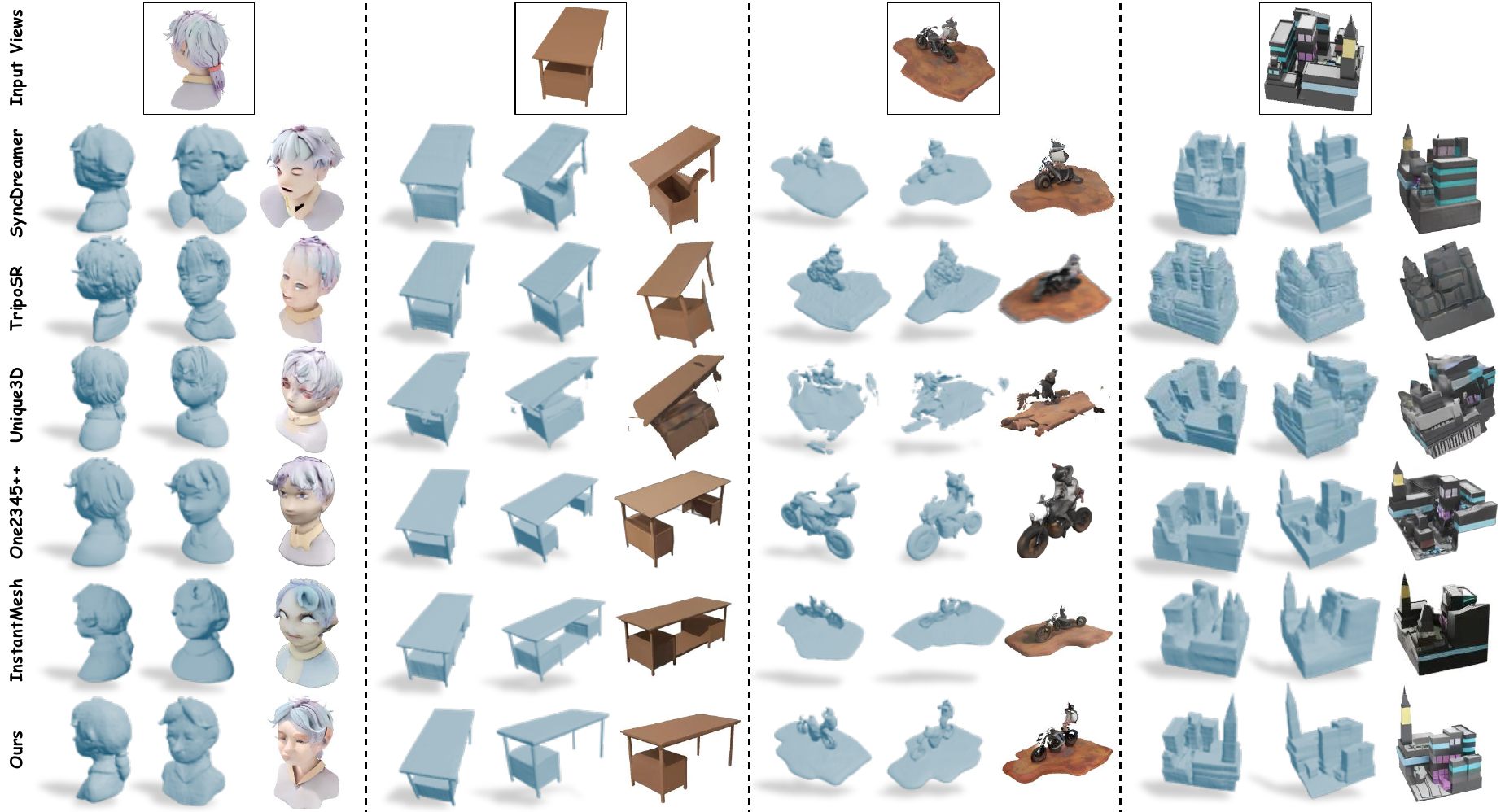}
    \vspace{-7pt}
    \caption{Visual comparisons between our \nMethod{} and recent cutting-edge methods on single-view image to 3D object generation. Note that the 3D results of Unique3D, TripoSR, and One2345++ are obtained by sending the input views to their official demos on Huggingface.}
    \label{fig:ito3d}
    \vspace{-7pt}
\end{figure*}
\subsection{Qualitative Results} \label{subsec:quali}

We perform qualitative experiments on novel view synthesis and image-to-3d on a wide range of 3D objects.
To highlight the advantages of our \nMethod{} method in contextual reasoning and zero-shot generalization, we adopt input views with a certain degree of offset relative to the frontal views of 3D samples.
We also provide more visualization results in the supplementary material.

\myPara{Novel View Synthesis.}
\figref{fig:view_syn} shows the synthesized views of our \nMethod{} and recent popular methods in multi-view generation, including Zero123-XL~\cite{liu2023zero}, SyncDreamer~\cite{liu2023syncdreamer}, Zero123++~\cite{shi2023zero123++}, One-2-3-45~\cite{liu2024one}.
Note that Zero123-XL is an enhanced Zero123 model pre-trained on the Objaverse-XL dataset, and the open-source project of One-2-3-45 also employs this Zero123 version to generate the 8 views for its first stage.
We utilize the elevation estimation implements of One-2-3-45 for the necessary elevation estimation procedures in Zero123-XL and SyncDreamer.
Among these difficult-to-maintain consistency scenes, some methods produce multiple inconsistent novel views, as shown in the bench results of Zero123++ and the cartoon figure of Zero123-XL. 
Some approaches may even be confused and generate new views that differ significantly from the input image, as shown in the bench predictions of SyncDreamer and the four-wheeled beds of One-2-3-45.
In contrast, our \nMethod{} is able to capture texture details of 3D objects and synthesize consistent multi-view images, which can be attributed to the full utilization of contextual information.

\myPara{Image-to-3D.}
Based on the synthesis of more consistent multiple views, our \nMethod{} can further generates high-quality 3D objects, as shown in \figref{fig:show}.
We also compare it with five cutting edge image-to-3d approaches, \ie SyncDreamer~\cite{liu2023syncdreamer}, InstantMesh~\cite{xu2024instantmesh}, One2345++~\cite{liu2024oneplus}, TripoSR~\cite{tochilkin2024triposr}, and Unique3D~\cite{wu2024unique3d}.
Note that the visual comparisons contain the pure geometries (left) and textured renderings (right) for each mesh generated by these methods.
As depicted in \figref{fig:ito3d}, our \nMethod{} can generate 3D meshes with a consistent appearance and plausible geometry under limited input view information.
Nevertheless, it is a real struggle for the counterparts to achieve this.
For example, InstantMesh and One2345++ tend to generate an additional storage cabinet for the desk. 
We speculate that this is due to their excessive reliance on the symmetry prior of diffusion models during the generative process of new views, with less consideration for the contextual information of the object itself.
Although SyncDreamer does not generate additional components, the desk it produces exhibits significant geometric deformations.
We find that the reason stems from the inconsistency among the 16 views generated by its diffusion model, which is employed to reconstruct the 3D object.
Unlike these approaches, our \nMethod{} effectively utilizes the contextual information of the object itself from nearby to distant during the autoregressive generation of all target views.
As a result, our method can achieve excellent performance in image-to-3D generation.

\subsection{Quantitative Results}

\begin{table*}[htp!]
    \setlength\tabcolsep{8pt}
    \centering
    \footnotesize
    \caption{
    Quantitative comparisons of our \nMethod{} model with the cutting-edge image-to-3d methods, including three 2D visual quality metrics and 3D geometric quality metrics.
    `$\uparrow$': the higher the value, the better the performance, `$\downarrow$': the lower the better.
    } 
  \vspace{-7pt}
  \label{tab:quan}
  \begin{tabular}{lcccccccccc} 
        \toprule
        \multirow{2}{*}{Methods} & \multicolumn{5}{c}{\makecell{GSO Dataset}} & \multicolumn{5}{c}{\makecell{Omni3D Dataset}}\\  \cmidrule(lr){2-6}  \cmidrule(lr){7-11}
        & \makecell{PSNR~$\uparrow$}& \makecell{SSIM~$\uparrow$}& \makecell{LPIPS~$\downarrow$}& \makecell{CD~$\downarrow$} & \makecell{F-Score~$\uparrow$} & \makecell{PSNR~$\uparrow$}& \makecell{SSIM~$\uparrow$}& \makecell{LPIPS~$\downarrow$}& \makecell{CD~$\downarrow$} &\makecell{F-Score~$\uparrow$} \\ 
        \midrule
        Michelangelo~\cite{zhao2023michelangelo} & 9.323 & 0.609 & 0.408 & 0.165 & 0.105 & 9.969 & 0.602 & 0.410 & 0.174 & 0.081 \\
        SyncDreamer~\cite{liu2023syncdreamer} &  10.82 & 0.652 & 0.332 & 0.108 & 0.125 & 9.485 & 0.585 & 0.436 & 0.196 & 0.067\\
        LGM~\cite{tang2024lgm} & 9.139 & 0.592 & 0.429 & 0.157 & 0.075 & 10.02 & 0.588 & 0.394 & 0.152 & 0.086\\
        InstantMesh~\cite{xu2024instantmesh} & 10.67 & 0.661 & 0.338 & 0.117 & 0.135 & 9.91 & 0.608 & 0.412 & 0.178 & 0.076 \\
        \midrule
        \nMethod{} (Ours) & \textbf{13.18} & \textbf{0.709} & \textbf{0.232} & \textbf{0.063} & \textbf{0.258} & \textbf{10.25} & \textbf{0.629} & \textbf{0.388} & \textbf{0.148} &\textbf{0.097} \\
        \bottomrule
  \end{tabular}
  \vspace{-10pt}
\end{table*}

We conduct quantitative studies on the two out-of-domain datasets, \ie GSO and Omni3D, to evaluate the performance of our \nMethod{} and other \sArt methods fairly.
Specifically, all candidate approaches are fed the same input images to generate 3D assets.
We render 20 views with 224$\times$224 resolution for each mesh to perform the 2D evaluation.
As far as the 3D evaluation, we uniformly sample 16K points from the mesh surfaces in an aligned
cube coordinate system of [-1,1]$^3$.
As shown in \tabref{tab:quan}, our \nMethod{} surpasses other cutting-edge methods across all metrics.
These results further demonstrate the superiority of our \nMethod{} in 3D asset creation.


\subsection{Ablative Study} \label{subsec:abl}

We conduct ablative experiments on the 300 objects excluded from training to examine the effectiveness of our key designs in the proposed \nMethod{} framework.

\begin{table}[tp]
    \setlength\tabcolsep{6.6pt}
    \centering
    \footnotesize
    \caption{Ablative study on conditional components of \nMethod{}.}
    \vspace{-7pt}
    \begin{tabular}{cccccc} 
      \toprule
      \nameoflocalmodule{} & \nameofglobalmodule{} & PSNR~$\uparrow$  & LPIPS~$\downarrow$ & SSIM~$\uparrow$  \\ \midrule
      & & 14.83 & 0.201 & 0.815 \\
      \CheckmarkBold & & 17.59 & 0.174 & 0.833 & \\
      & \CheckmarkBold & 17.91 & 0.170 & 0.836 & \\
      \CheckmarkBold & \CheckmarkBold & 20.28 & 0.121 & 0.857 \\
      \bottomrule
    \end{tabular}
    \vspace{-5pt}
    \label{tab:abl_components}
\end{table}

\begin{table}[tp]
    \setlength\tabcolsep{6pt}
    \centering
    \footnotesize
    \caption{Ablative study on sequence orders during the autoregressive generation of \nMethod{}.} 
    \vspace{-7pt}
      \begin{tabular}{lcccc} 
          \toprule
          Model & PSNR~$\uparrow$  & LPIPS~$\downarrow$ & SSIM~$\uparrow$ & CLIP-Score~$\uparrow$ \\ \midrule
          Reverse & 20.19 & 0.124 & 0.851 & 0.882 \\
          Random & 17.36 & 0.167 & 0.839 & 0.774 \\
          Normal (Ours) & 20.28 & 0.121 & 0.857 & 0.887 \\
          \bottomrule
      \end{tabular}
    \vspace{-12pt}
    \label{tab:abl_order}
\end{table}

\myPara{Ablation on Image Conditioning Components.}
Starting from the Zero123++ baseline, we first incorporate \nameoflocalmodule{} strategy, then \nameofglobalmodule{} independently, and finally integrate both strategies.
As shown in \tabref{tab:abl_components}, they individually contribute to performance improvements, and their combined integration produces a greater improvement compared to the baseline method.
These results demonstrate both the effectiveness of the two image conditioning strategies and the significant contribution of our next-view prediction in advancing the accuracy of novel view synthesis.

\myPara{Ablation on Different Sequence Orders.}
We define the sequence order from near to far in terms of camera poses relative to the input viewpoint as the normal order.
We also provide two variants of the sequence order, \ie reverse order and random order, to generate target views.
To be specific, the reverse order refers to the view sequence whose cameras move from far to near relative to the input view.
Meanwhile, the random order places the middle row of the $3 \times 2$ grid layout as the first position, followed by the remaining two rows.
As shown in \tabref{tab:abl_order}, our normal order achieves the best performance, while the random sequence performs the worst, demonstrating the effectiveness of modeling the target views in a sequential manner.
Note that the reverse order achieves a similar performance to our normal order.
We believe the reason for this is that the reversed sequence under Zero123++ setting can be seen as the camera moving from near to far in another circle direction, as the camera moves in a circular motion around the 3D object.

\begin{figure}[t]
    \centering
    \footnotesize
    \vspace{2pt}
    \includegraphics[width=0.98\linewidth]{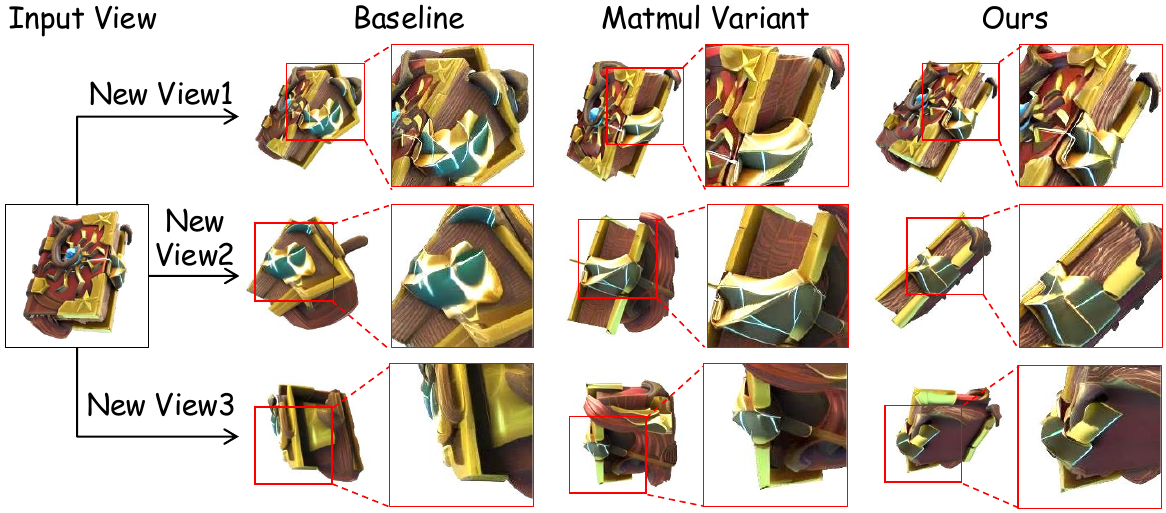}
    \vspace{-7pt}
    \caption{Ablative studies on the global feature encoding strategy.
    }
    \label{fig:abl_enc}
    \vspace{-10pt}
\end{figure}

\myPara{Encoding Strategy of Global Feature Sequence.}
%
To highlight the effectiveness of our \nameofglobalmodule{} for global feature encoding of view sequence, we design a matrix multiplication (`matmul') variant to encode these features.
Specifically, we stack these features into a matrix with shape $\mathbb{R}^{(2k-1) \times D}$.
Meanwhile, we repeat the trainable weights in global condition $(2k$-$1)$ times to obtain a matrix with shape $\mathbb{R}^{77 \times (2k-1)}$.
We multiply these two matrices to obtain a new matrix with shape $\mathbb{R}^{77 \times D}$, which is utilized as the key and value matrix for the cross-attention mechanism of the denoising UNet.
As depicted in \figref{fig:abl_enc}, with the incorporation of extra contextual information, the `matmul' variant can generate more consistent and high-quality multi-view images compared to the baseline method, \ie Zero123++.  
Nevertheless, this variant may lead to bias in the global semantic understanding of 3D objects, such as the shapes of the book sample.
In contrast, our strategy can generate multi-view images that are faithful to the shape and texture of objects in the input views.
These experiments indicate the LSTM proposal can more
effectively capture the high-level semantic information of the 3D objects.

\section{Conclusions}
In this paper, we present \nMethod{}, a next-view prediction paradigm that starts from the input image and progressively generates target views from near to far.
At each step of the autoregressive process, the previously generated views are employed as contextual information to facilitate the generation of the current target views.
The experimental results demonstrate that our method generates new view images and 3D objects that are more consistent with the input images compared with the existing approaches generating multi-view images discretely or simultaneously.

\myPara{Acknowledgments.}
This research was partially funded by Shenzhen Science and Technology Program (No. JCYJ20240813114237048).
Besides, we express our gratitude to Dr. Xin Chen for his assistance in revising the paper.

{
    \small
    \bibliographystyle{ieeenat_fullname}
    \bibliography{main}

\begin{thebibliography}{77}
\providecommand{\natexlab}[1]{#1}
\providecommand{\url}[1]{\texttt{#1}}
\expandafter\ifx\csname urlstyle\endcsname\relax
  \providecommand{\doi}[1]{doi: #1}\else
  \providecommand{\doi}{doi: \begingroup \urlstyle{rm}\Url}\fi

\bibitem[Chen et~al.(2025)Chen, Mart{\'\i}~Mons{\'o}, Du, Simchowitz, Tedrake, and Sitzmann]{chen2025diffusion}
Boyuan Chen, Diego Mart{\'\i}~Mons{\'o}, Yilun Du, Max Simchowitz, Russ Tedrake, and Vincent Sitzmann.
\newblock Diffusion forcing: Next-token prediction meets full-sequence diffusion.
\newblock \emph{Advances in Neural Information Processing Systems}, 37:\penalty0 24081--24125, 2025.

\bibitem[Chen et~al.(2023{\natexlab{a}})Chen, Chen, Jiao, and Jia]{chen2023fantasia3d}
Rui Chen, Yongwei Chen, Ningxin Jiao, and Kui Jia.
\newblock Fantasia3d: Disentangling geometry and appearance for high-quality text-to-3d content creation.
\newblock In \emph{Proceedings of the IEEE/CVF international conference on computer vision}, pages 22246--22256, 2023{\natexlab{a}}.

\bibitem[Chen et~al.(2024{\natexlab{a}})Chen, Chen, Pang, Zeng, Cheng, Fu, Yin, Wang, Wang, Zhang, et~al.]{chen2024meshxl}
Sijin Chen, Xin Chen, Anqi Pang, Xianfang Zeng, Wei Cheng, Yijun Fu, Fukun Yin, Yanru Wang, Zhibin Wang, Chi Zhang, et~al.
\newblock Meshxl: Neural coordinate field for generative 3d foundation models.
\newblock \emph{arXiv preprint arXiv:2405.20853}, 2024{\natexlab{a}}.

\bibitem[Chen et~al.(2023{\natexlab{b}})Chen, Fang, Huang, Yi, Zhang, Xie, Wang, Dai, Xiong, and Tian]{chen2023cascade}
Yabo Chen, Jiemin Fang, Yuyang Huang, Taoran Yi, Xiaopeng Zhang, Lingxi Xie, Xinggang Wang, Wenrui Dai, Hongkai Xiong, and Qi Tian.
\newblock Cascade-zero123: One image to highly consistent 3d with self-prompted nearby views.
\newblock \emph{arXiv preprint arXiv:2312.04424}, 2023{\natexlab{b}}.

\bibitem[Chen et~al.(2024{\natexlab{b}})Chen, He, Huang, Ye, Chen, Tang, Chen, Cai, Yang, Yu, et~al.]{chen2024meshanything}
Yiwen Chen, Tong He, Di Huang, Weicai Ye, Sijin Chen, Jiaxiang Tang, Xin Chen, Zhongang Cai, Lei Yang, Gang Yu, et~al.
\newblock Meshanything: Artist-created mesh generation with autoregressive transformers.
\newblock \emph{arXiv preprint arXiv:2406.10163}, 2024{\natexlab{b}}.

\bibitem[Chen et~al.(2024{\natexlab{c}})Chen, Lan, Zhou, Wang, and Pan]{chen2024sar3d}
Yongwei Chen, Yushi Lan, Shangchen Zhou, Tengfei Wang, and XIngang Pan.
\newblock Sar3d: Autoregressive 3d object generation and understanding via multi-scale 3d vqvae.
\newblock \emph{arXiv preprint arXiv:2411.16856}, 2024{\natexlab{c}}.

\bibitem[Chen et~al.(2024{\natexlab{d}})Chen, Li, Wang, Chen, Jiang, Li, Wang, Yang, and Tai]{chen2024region}
Zhennan Chen, Yajie Li, Haofan Wang, Zhibo Chen, Zhengkai Jiang, Jun Li, Qian Wang, Jian Yang, and Ying Tai.
\newblock Region-aware text-to-image generation via hard binding and soft refinement.
\newblock \emph{arXiv preprint arXiv:2411.06558}, 2024{\natexlab{d}}.

\bibitem[Chen et~al.(2024{\natexlab{e}})Chen, Zhang, Xiang, and Tai]{chen2024adaptive}
Zhennan Chen, Xuying Zhang, Tian-Zhu Xiang, and Ying Tai.
\newblock Adaptive guidance learning for camouflaged object detection.
\newblock \emph{arXiv preprint arXiv:2405.02824}, 2024{\natexlab{e}}.

\bibitem[Deitke et~al.(2023)Deitke, Schwenk, Salvador, Weihs, Michel, VanderBilt, Schmidt, Ehsani, Kembhavi, and Farhadi]{deitke2023objaverse}
Matt Deitke, Dustin Schwenk, Jordi Salvador, Luca Weihs, Oscar Michel, Eli VanderBilt, Ludwig Schmidt, Kiana Ehsani, Aniruddha Kembhavi, and Ali Farhadi.
\newblock Objaverse: A universe of annotated 3d objects.
\newblock In \emph{Proceedings of the IEEE/CVF Conference on Computer Vision and Pattern Recognition}, 2023.

\bibitem[Downs et~al.(2022)Downs, Francis, Koenig, Kinman, Hickman, Reymann, McHugh, and Vanhoucke]{downs2022google}
Laura Downs, Anthony Francis, Nate Koenig, Brandon Kinman, Ryan Hickman, Krista Reymann, Thomas~B McHugh, and Vincent Vanhoucke.
\newblock Google scanned objects: A high-quality dataset of 3d scanned household items.
\newblock In \emph{2022 International Conference on Robotics and Automation (ICRA)}, pages 2553--2560. IEEE, 2022.

\bibitem[Du et~al.(2025)Du, Chen, Chen, Gao, Chen, Jiang, Yang, and Tai]{du2025textcrafter}
Nikai Du, Zhennan Chen, Zhizhou Chen, Shan Gao, Xi Chen, Zhengkai Jiang, Jian Yang, and Ying Tai.
\newblock Textcrafter: Accurately rendering multiple texts in complex visual scenes.
\newblock \emph{arXiv preprint arXiv:2503.23461}, 2025.

\bibitem[Esser et~al.(2021)Esser, Rombach, and Ommer]{esser2021taming}
Patrick Esser, Robin Rombach, and Bjorn Ommer.
\newblock Taming transformers for high-resolution image synthesis.
\newblock In \emph{Proceedings of the IEEE/CVF conference on computer vision and pattern recognition}, pages 12873--12883, 2021.

\bibitem[Gao et~al.(2024)Gao, Holynski, Henzler, Brussee, Martin-Brualla, Srinivasan, Barron, and Poole]{gao2024cat3d}
Ruiqi Gao, Aleksander Holynski, Philipp Henzler, Arthur Brussee, Ricardo Martin-Brualla, Pratul Srinivasan, Jonathan~T Barron, and Ben Poole.
\newblock Cat3d: Create anything in 3d with multi-view diffusion models.
\newblock \emph{arXiv preprint arXiv:2405.10314}, 2024.

\bibitem[Han et~al.(2024)Han, Liu, Jiang, Yan, Zhang, Yuan, Peng, and Liu]{han2024infinity}
Jian Han, Jinlai Liu, Yi Jiang, Bin Yan, Yuqi Zhang, Zehuan Yuan, Bingyue Peng, and Xiaobing Liu.
\newblock Infinity: Scaling bitwise autoregressive modeling for high-resolution image synthesis.
\newblock \emph{arXiv preprint arXiv:2412.04431}, 2024.

\bibitem[Hochreiter(1997)]{hochreiter1997long}
S Hochreiter.
\newblock Long short-term memory.
\newblock \emph{Neural Computation MIT-Press}, 1997.

\bibitem[Hong et~al.(2023)Hong, Zhang, Gu, Bi, Zhou, Liu, Liu, Sunkavalli, Bui, and Tan]{hong2023lrm}
Yicong Hong, Kai Zhang, Jiuxiang Gu, Sai Bi, Yang Zhou, Difan Liu, Feng Liu, Kalyan Sunkavalli, Trung Bui, and Hao Tan.
\newblock Lrm: Large reconstruction model for single image to 3d.
\newblock \emph{arXiv preprint arXiv:2311.04400}, 2023.

\bibitem[Hu et~al.(2024)Hu, Zhou, Jampani, and Tulsiani]{hu2024mvd}
Hanzhe Hu, Zhizhuo Zhou, Varun Jampani, and Shubham Tulsiani.
\newblock Mvd-fusion: Single-view 3d via depth-consistent multi-view generation.
\newblock In \emph{Proceedings of the IEEE/CVF Conference on Computer Vision and Pattern Recognition}, pages 9698--9707, 2024.

\bibitem[Ji et~al.(2024)Ji, Wang, Wu, Ma, Sun, and Ji]{ji2024jm3d}
Jiayi Ji, Haowei Wang, Changli Wu, Yiwei Ma, Xiaoshuai Sun, and Rongrong Ji.
\newblock Jm3d \& jm3d-llm: Elevating 3d representation with joint multi-modal cues.
\newblock \emph{IEEE Transactions on Pattern Analysis and Machine Intelligence}, 2024.

\bibitem[Jin et~al.(2024)Jin, Sun, Li, Xu, Jiang, Zhuang, Huang, Song, Mu, and Lin]{jin2024pyramidal}
Yang Jin, Zhicheng Sun, Ningyuan Li, Kun Xu, Hao Jiang, Nan Zhuang, Quzhe Huang, Yang Song, Yadong Mu, and Zhouchen Lin.
\newblock Pyramidal flow matching for efficient video generative modeling.
\newblock \emph{arXiv preprint arXiv:2410.05954}, 2024.

\bibitem[Labs(2024)]{black2024flux}
Black~Forest Labs.
\newblock Flux.
\newblock \emph{https://github.com/black-forest-labs/flux}, 2024.

\bibitem[Li et~al.(2023)Li, Tan, Zhang, Xu, Luan, Xu, Hong, Sunkavalli, Shakhnarovich, and Bi]{li2023instant3d}
Jiahao Li, Hao Tan, Kai Zhang, Zexiang Xu, Fujun Luan, Yinghao Xu, Yicong Hong, Kalyan Sunkavalli, Greg Shakhnarovich, and Sai Bi.
\newblock Instant3d: Fast text-to-3d with sparse-view generation and large reconstruction model.
\newblock \emph{arXiv preprint arXiv:2311.06214}, 2023.

\bibitem[Li et~al.(2024{\natexlab{a}})Li, Lai, Xu, Guo, Cao, Zhang, Dai, and Ji]{li2024dual3d}
Xinyang Li, Zhangyu Lai, Linning Xu, Jianfei Guo, Liujuan Cao, Shengchuan Zhang, Bo Dai, and Rongrong Ji.
\newblock Dual3d: Efficient and consistent text-to-3d generation with dual-mode multi-view latent diffusion.
\newblock \emph{arXiv preprint arXiv:2405.09874}, 2024{\natexlab{a}}.

\bibitem[Li et~al.(2024{\natexlab{b}})Li, Lai, Xu, Qu, Cao, Zhang, Dai, and Ji]{li2024director3d}
Xinyang Li, Zhangyu Lai, Linning Xu, Yansong Qu, Liujuan Cao, Shengchuan Zhang, Bo Dai, and Rongrong Ji.
\newblock Director3d: Real-world camera trajectory and 3d scene generation from text.
\newblock \emph{Advances in neural information processing systems}, 37:\penalty0 75125--75151, 2024{\natexlab{b}}.

\bibitem[Li et~al.(2025)Li, Zou, Liu, Wang, Liang, Yu, Liu, Guo, Liang, Ouyang, et~al.]{li2025triposg}
Yangguang Li, Zi-Xin Zou, Zexiang Liu, Dehu Wang, Yuan Liang, Zhipeng Yu, Xingchao Liu, Yuan-Chen Guo, Ding Liang, Wanli Ouyang, et~al.
\newblock Triposg: High-fidelity 3d shape synthesis using large-scale rectified flow models.
\newblock \emph{arXiv preprint arXiv:2502.06608}, 2025.

\bibitem[Li et~al.(2024{\natexlab{c}})Li, Cao, Wang, Qi, Cheng, and Shan]{li2024photomaker}
Zhen Li, Mingdeng Cao, Xintao Wang, Zhongang Qi, Ming-Ming Cheng, and Ying Shan.
\newblock Photomaker: Customizing realistic human photos via stacked id embedding.
\newblock In \emph{Proceedings of the IEEE/CVF Conference on Computer Vision and Pattern Recognition}, pages 8640--8650, 2024{\natexlab{c}}.

\bibitem[Li et~al.(2024{\natexlab{d}})Li, Song, Zhao, Cheng, Li, and Yang]{li2024atprompt}
Zheng Li, Yibing Song, Penghai Zhao, Ming-Ming Cheng, Xiang Li, and Jian Yang.
\newblock Atprompt: Textual prompt learning with embedded attributes.
\newblock \emph{arXiv preprint arXiv:2412.09442}, 2024{\natexlab{d}}.

\bibitem[Lin et~al.(2023)Lin, Han, Gong, Xu, Zhang, and Li]{lin2023consistent123}
Yukang Lin, Haonan Han, Chaoqun Gong, Zunnan Xu, Yachao Zhang, and Xiu Li.
\newblock Consistent123: One image to highly consistent 3d asset using case-aware diffusion priors.
\newblock \emph{arXiv preprint arXiv:2309.17261}, 2023.

\bibitem[Liu et~al.(2024{\natexlab{a}})Liu, Shi, Chen, Zhang, Xu, Wei, Chen, Zeng, Gu, and Su]{liu2024oneplus}
Minghua Liu, Ruoxi Shi, Linghao Chen, Zhuoyang Zhang, Chao Xu, Xinyue Wei, Hansheng Chen, Chong Zeng, Jiayuan Gu, and Hao Su.
\newblock One-2-3-45++: Fast single image to 3d objects with consistent multi-view generation and 3d diffusion.
\newblock In \emph{Proceedings of the IEEE/CVF Conference on Computer Vision and Pattern Recognition}, pages 10072--10083, 2024{\natexlab{a}}.

\bibitem[Liu et~al.(2024{\natexlab{b}})Liu, Xu, Jin, Chen, Varma~T, Xu, and Su]{liu2024one}
Minghua Liu, Chao Xu, Haian Jin, Linghao Chen, Mukund Varma~T, Zexiang Xu, and Hao Su.
\newblock One-2-3-45: Any single image to 3d mesh in 45 seconds without per-shape optimization.
\newblock \emph{Advances in Neural Information Processing Systems}, 36, 2024{\natexlab{b}}.

\bibitem[Liu et~al.(2023{\natexlab{a}})Liu, Wu, Van~Hoorick, Tokmakov, Zakharov, and Vondrick]{liu2023zero}
Ruoshi Liu, Rundi Wu, Basile Van~Hoorick, Pavel Tokmakov, Sergey Zakharov, and Carl Vondrick.
\newblock Zero-1-to-3: Zero-shot one image to 3d object.
\newblock In \emph{Proceedings of the IEEE/CVF international conference on computer vision}, pages 9298--9309, 2023{\natexlab{a}}.

\bibitem[Liu et~al.(2023{\natexlab{b}})Liu, Lin, Zeng, Long, Liu, Komura, and Wang]{liu2023syncdreamer}
Yuan Liu, Cheng Lin, Zijiao Zeng, Xiaoxiao Long, Lingjie Liu, Taku Komura, and Wenping Wang.
\newblock Syncdreamer: Generating multiview-consistent images from a single-view image.
\newblock \emph{arXiv preprint arXiv:2309.03453}, 2023{\natexlab{b}}.

\bibitem[Long et~al.(2024)Long, Guo, Lin, Liu, Dou, Liu, Ma, Zhang, Habermann, Theobalt, et~al.]{long2024wonder3d}
Xiaoxiao Long, Yuan-Chen Guo, Cheng Lin, Yuan Liu, Zhiyang Dou, Lingjie Liu, Yuexin Ma, Song-Hai Zhang, Marc Habermann, Christian Theobalt, et~al.
\newblock Wonder3d: Single image to 3d using cross-domain diffusion.
\newblock In \emph{Proceedings of the IEEE/CVF Conference on Computer Vision and Pattern Recognition}, pages 9970--9980, 2024.

\bibitem[Loshchilov and Hutter(2016)]{loshchilov2016sgdr}
Ilya Loshchilov and Frank Hutter.
\newblock Sgdr: Stochastic gradient descent with warm restarts.
\newblock \emph{arXiv preprint arXiv:1608.03983}, 2016.

\bibitem[Loshchilov and Hutter(2019)]{loshchilov2019decoupled}
Ilya Loshchilov and Frank Hutter.
\newblock Decoupled weight decay regularization.
\newblock In \emph{ICLR}, 2019.

\bibitem[Lu et~al.(2025)Lu, Ji, Zhu, and Cao]{lu2025learning}
Yuanxun Lu, Xinya Ji, Hao Zhu, and Xun Cao.
\newblock Learning hierarchical adaptive code clouds for neural 3d shape representation.
\newblock \emph{Machine Intelligence Research}, 22\penalty0 (2):\penalty0 304--323, 2025.

\bibitem[Ma et~al.(2023)Ma, Ji, Sun, Zhou, and Ji]{ma2023towards}
Yiwei Ma, Jiayi Ji, Xiaoshuai Sun, Yiyi Zhou, and Rongrong Ji.
\newblock Towards local visual modeling for image captioning.
\newblock \emph{Pattern Recognition}, 138:\penalty0 109420, 2023.

\bibitem[Mu et~al.(2023)Mu, Chen, Cai, and Guo]{mu2023neural}
Tai-Jiang Mu, Hao-Xiang Chen, Jun-Xiong Cai, and Ning Guo.
\newblock Neural 3d reconstruction from sparse views using geometric priors.
\newblock \emph{Computational Visual Media}, 9\penalty0 (4):\penalty0 687--697, 2023.

\bibitem[Peebles and Xie(2023)]{peebles2023scalable}
William Peebles and Saining Xie.
\newblock Scalable diffusion models with transformers.
\newblock In \emph{Proceedings of the IEEE/CVF International Conference on Computer Vision}, pages 4195--4205, 2023.

\bibitem[Poole et~al.(2022)Poole, Jain, Barron, and Mildenhall]{poole2022dreamfusion}
Ben Poole, Ajay Jain, Jonathan~T Barron, and Ben Mildenhall.
\newblock Dreamfusion: Text-to-3d using 2d diffusion.
\newblock \emph{arXiv preprint arXiv:2209.14988}, 2022.

\bibitem[Qian et~al.(2023)Qian, Mai, Hamdi, Ren, Siarohin, Li, Lee, Skorokhodov, Wonka, Tulyakov, et~al.]{qian2023magic123}
Guocheng Qian, Jinjie Mai, Abdullah Hamdi, Jian Ren, Aliaksandr Siarohin, Bing Li, Hsin-Ying Lee, Ivan Skorokhodov, Peter Wonka, Sergey Tulyakov, et~al.
\newblock Magic123: One image to high-quality 3d object generation using both 2d and 3d diffusion priors.
\newblock \emph{arXiv preprint arXiv:2306.17843}, 2023.

\bibitem[Qu et~al.(2025)Qu, Chen, Li, Li, Zhang, Cao, and Ji]{qu2025drag}
Yansong Qu, Dian Chen, Xinyang Li, Xiaofan Li, Shengchuan Zhang, Liujuan Cao, and Rongrong Ji.
\newblock Drag your gaussian: Effective drag-based editing with score distillation for 3d gaussian splatting.
\newblock \emph{arXiv preprint arXiv:2501.18672}, 2025.

\bibitem[Radford et~al.(2021)Radford, Kim, Hallacy, Ramesh, Goh, Agarwal, Sastry, Askell, Mishkin, Clark, et~al.]{radford2021learning}
Alec Radford, Jong~Wook Kim, Chris Hallacy, Aditya Ramesh, Gabriel Goh, Sandhini Agarwal, Girish Sastry, Amanda Askell, Pamela Mishkin, Jack Clark, et~al.
\newblock Learning transferable visual models from natural language supervision.
\newblock In \emph{International conference on machine learning}. PMLR, 2021.

\bibitem[Rombach et~al.(2022)Rombach, Blattmann, Lorenz, Esser, and Ommer]{rombach2022high}
Robin Rombach, Andreas Blattmann, Dominik Lorenz, Patrick Esser, and Bj{\"o}rn Ommer.
\newblock High-resolution image synthesis with latent diffusion models.
\newblock In \emph{CVPR}, 2022.

\bibitem[Ruhe et~al.(2024)Ruhe, Heek, Salimans, and Hoogeboom]{ruhe2024rolling}
David Ruhe, Jonathan Heek, Tim Salimans, and Emiel Hoogeboom.
\newblock Rolling diffusion models.
\newblock \emph{arXiv preprint arXiv:2402.09470}, 2024.

\bibitem[Shi et~al.(2023{\natexlab{a}})Shi, Chen, Zhang, Liu, Xu, Wei, Chen, Zeng, and Su]{shi2023zero123++}
Ruoxi Shi, Hansheng Chen, Zhuoyang Zhang, Minghua Liu, Chao Xu, Xinyue Wei, Linghao Chen, Chong Zeng, and Hao Su.
\newblock Zero123++: a single image to consistent multi-view diffusion base model.
\newblock \emph{arXiv preprint arXiv:2310.15110}, 2023{\natexlab{a}}.

\bibitem[Shi et~al.(2015)Shi, Chen, Wang, Yeung, Wong, and Woo]{shi2015convolutional}
Xingjian Shi, Zhourong Chen, Hao Wang, Dit-Yan Yeung, Wai-Kin Wong, and Wang-chun Woo.
\newblock Convolutional lstm network: A machine learning approach for precipitation nowcasting.
\newblock \emph{Advances in neural information processing systems}, 28, 2015.

\bibitem[Shi et~al.(2023{\natexlab{b}})Shi, Wang, Ye, Long, Li, and Yang]{shi2023mvdream}
Yichun Shi, Peng Wang, Jianglong Ye, Mai Long, Kejie Li, and Xiao Yang.
\newblock Mvdream: Multi-view diffusion for 3d generation.
\newblock \emph{arXiv preprint}, 2023{\natexlab{b}}.

\bibitem[Siddiqui et~al.(2024)Siddiqui, Alliegro, Artemov, Tommasi, Sirigatti, Rosov, Dai, and Nie{\ss}ner]{siddiqui2024meshgpt}
Yawar Siddiqui, Antonio Alliegro, Alexey Artemov, Tatiana Tommasi, Daniele Sirigatti, Vladislav Rosov, Angela Dai, and Matthias Nie{\ss}ner.
\newblock Meshgpt: Generating triangle meshes with decoder-only transformers.
\newblock In \emph{Proceedings of the IEEE/CVF Conference on Computer Vision and Pattern Recognition}, pages 19615--19625, 2024.

\bibitem[Sun et~al.(2024)Sun, Jiang, Chen, Zhang, Peng, Luo, and Yuan]{sun2024autoregressive}
Peize Sun, Yi Jiang, Shoufa Chen, Shilong Zhang, Bingyue Peng, Ping Luo, and Zehuan Yuan.
\newblock Autoregressive model beats diffusion: Llama for scalable image generation.
\newblock \emph{arXiv preprint arXiv:2406.06525}, 2024.

\bibitem[Tang et~al.(2024{\natexlab{a}})Tang, Chen, Chen, Wang, Zeng, and Liu]{tang2024lgm}
Jiaxiang Tang, Zhaoxi Chen, Xiaokang Chen, Tengfei Wang, Gang Zeng, and Ziwei Liu.
\newblock Lgm: Large multi-view gaussian model for high-resolution 3d content creation.
\newblock In \emph{European Conference on Computer Vision}, pages 1--18. Springer, 2024{\natexlab{a}}.

\bibitem[Tang et~al.(2023)Tang, Zhang, Chen, Wang, and Furukawa]{Tang2023mvdiffusion}
Shitao Tang, Fuyang Zhang, Jiacheng Chen, Peng Wang, and Yasutaka Furukawa.
\newblock Mvdiffusion: Enabling holistic multi-view image generation with correspondence-aware diffusion.
\newblock \emph{arXiv}, 2023.

\bibitem[Tang et~al.(2024{\natexlab{b}})Tang, Zhang, Cheng, Yu, Feng, Pang, Lin, and Yuan]{tang2024cycle3d}
Zhenyu Tang, Junwu Zhang, Xinhua Cheng, Wangbo Yu, Chaoran Feng, Yatian Pang, Bin Lin, and Li Yuan.
\newblock Cycle3d: High-quality and consistent image-to-3d generation via generation-reconstruction cycle.
\newblock \emph{arXiv preprint arXiv:2407.19548}, 2024{\natexlab{b}}.

\bibitem[Tian et~al.(2024)Tian, Jiang, Yuan, Peng, and Wang]{tian2024visual}
Keyu Tian, Yi Jiang, Zehuan Yuan, Bingyue Peng, and Liwei Wang.
\newblock Visual autoregressive modeling: Scalable image generation via next-scale prediction.
\newblock \emph{arXiv preprint arXiv:2404.02905}, 2024.

\bibitem[Tochilkin et~al.(2024)Tochilkin, Pankratz, Liu, Huang, Letts, Li, Liang, Laforte, Jampani, and Cao]{tochilkin2024triposr}
Dmitry Tochilkin, David Pankratz, Zexiang Liu, Zixuan Huang, Adam Letts, Yangguang Li, Ding Liang, Christian Laforte, Varun Jampani, and Yan-Pei Cao.
\newblock Triposr: Fast 3d object reconstruction from a single image.
\newblock \emph{arXiv preprint arXiv:2403.02151}, 2024.

\bibitem[Van Den~Oord et~al.(2016)Van Den~Oord, Kalchbrenner, and Kavukcuoglu]{van2016pixel}
A{\"a}ron Van Den~Oord, Nal Kalchbrenner, and Koray Kavukcuoglu.
\newblock Pixel recurrent neural networks.
\newblock In \emph{International conference on machine learning}, pages 1747--1756. PMLR, 2016.

\bibitem[Van Den~Oord et~al.(2017)Van Den~Oord, Vinyals, et~al.]{van2017neural}
Aaron Van Den~Oord, Oriol Vinyals, et~al.
\newblock Neural discrete representation learning.
\newblock \emph{Advances in neural information processing systems}, 30, 2017.

\bibitem[Vaswani(2017)]{vaswani2017attention}
A Vaswani.
\newblock Attention is all you need.
\newblock \emph{Advances in Neural Information Processing Systems}, 2017.

\bibitem[Voleti et~al.(2024)Voleti, Yao, Boss, Letts, Pankratz, Tochilkin, Laforte, Rombach, and Jampani]{voleti2024sv3d}
Vikram Voleti, Chun-Han Yao, Mark Boss, Adam Letts, David Pankratz, Dmitry Tochilkin, Christian Laforte, Robin Rombach, and Varun Jampani.
\newblock Sv3d: Novel multi-view synthesis and 3d generation from a single image using latent video diffusion.
\newblock In \emph{European Conference on Computer Vision}, pages 439--457. Springer, 2024.

\bibitem[Wang et~al.(2023)Wang, Ji, Guo, Yang, Zhou, Sun, and Ji]{wang2023nice}
Haowei Wang, Jiayi Ji, Tianyu Guo, Yilong Yang, Yiyi Zhou, Xiaoshuai Sun, and Rongrong Ji.
\newblock Nice: improving panoptic narrative detection and segmentation with cascading collaborative learning.
\newblock \emph{arXiv preprint arXiv:2310.10975}, 2023.

\bibitem[Wang and Shi(2023)]{wang2023imagedream}
Peng Wang and Yichun Shi.
\newblock Imagedream: Image-prompt multi-view diffusion for 3d generation.
\newblock \emph{arXiv preprint arXiv:2312.02201}, 2023.

\bibitem[Wang et~al.(2004)Wang, Bovik, Sheikh, and Simoncelli]{wang2004image}
Zhou Wang, Alan~C Bovik, Hamid~R Sheikh, and Eero~P Simoncelli.
\newblock Image quality assessment: from error visibility to structural similarity.
\newblock \emph{IEEE transactions on image processing}, 13\penalty0 (4):\penalty0 600--612, 2004.

\bibitem[Wu et~al.(2024)Wu, Liu, Cai, Yan, Wang, Hu, Duan, and Ma]{wu2024unique3d}
Kailu Wu, Fangfu Liu, Zhihan Cai, Runjie Yan, Hanyang Wang, Yating Hu, Yueqi Duan, and Kaisheng Ma.
\newblock Unique3d: High-quality and efficient 3d mesh generation from a single image.
\newblock \emph{arXiv preprint arXiv:2405.20343}, 2024.

\bibitem[Wu et~al.(2023)Wu, Zhang, Fu, Wang, Ren, Pan, Wu, Yang, Wang, Qian, et~al.]{wu2023omniobject3d}
Tong Wu, Jiarui Zhang, Xiao Fu, Yuxin Wang, Jiawei Ren, Liang Pan, Wayne Wu, Lei Yang, Jiaqi Wang, Chen Qian, et~al.
\newblock Omniobject3d: Large-vocabulary 3d object dataset for realistic perception, reconstruction and generation.
\newblock In \emph{Proceedings of the IEEE/CVF Conference on Computer Vision and Pattern Recognition}, pages 803--814, 2023.

\bibitem[Wu et~al.(2025)Wu, Zhang, Liu, Zhang, and Cheng]{wu2025ret3d}
Yu-Huan Wu, Da Zhang, Yun Liu, Le Zhang, and Ming-Ming Cheng.
\newblock Ret3d: rethinking object relations for efficient 3d object detection.
\newblock \emph{Sci~Sin~Inform}, 55\penalty0 (4):\penalty0 887--901, 2025.

\bibitem[Xu et~al.(2024)Xu, Cheng, Gao, Wang, Gao, and Shan]{xu2024instantmesh}
Jiale Xu, Weihao Cheng, Yiming Gao, Xintao Wang, Shenghua Gao, and Ying Shan.
\newblock Instantmesh: Efficient 3d mesh generation from a single image with sparse-view large reconstruction models.
\newblock \emph{arXiv preprint arXiv:2404.07191}, 2024.

\bibitem[Yu et~al.(2022)Yu, Xu, Koh, Luong, Baid, Wang, Vasudevan, Ku, Yang, Ayan, et~al.]{yu2022scaling}
Jiahui Yu, Yuanzhong Xu, Jing~Yu Koh, Thang Luong, Gunjan Baid, Zirui Wang, Vijay Vasudevan, Alexander Ku, Yinfei Yang, Burcu~Karagol Ayan, et~al.
\newblock Scaling autoregressive models for content-rich text-to-image generation.
\newblock \emph{arXiv preprint arXiv:2206.10789}, 2\penalty0 (3):\penalty0 5, 2022.

\bibitem[Zhang(2023)]{zhang2023reference}
Lyumin Zhang.
\newblock Reference-only control.
\newblock \emph{https:// github.com/Mikubill/sd-webui-controlnet/ discussions/1236}, 2023.

\bibitem[Zhang et~al.(2018)Zhang, Isola, Efros, Shechtman, and Wang]{zhang2018unreasonable}
Richard Zhang, Phillip Isola, Alexei~A Efros, Eli Shechtman, and Oliver Wang.
\newblock The unreasonable effectiveness of deep features as a perceptual metric.
\newblock In \emph{Proceedings of the IEEE conference on computer vision and pattern recognition}, pages 586--595, 2018.

\bibitem[Zhang et~al.(2024{\natexlab{a}})Zhang, Qian, Xie, and Jian]{zhang24FastPCI}
Tianyu Zhang, Guocheng Qian, Jin Xie, and Yang Jian.
\newblock Fastpci: Motion-structure guided fast point cloud frame interpolation.
\newblock In \emph{ECCV}, 2024{\natexlab{a}}.

\bibitem[Zhang et~al.(2024{\natexlab{b}})Zhang, Liu, Li, Zhang, Liu, Wang, Ouyang, Xiong, Gao, Hou, et~al.]{zhang2024tar3d}
Xuying Zhang, Yutong Liu, Yangguang Li, Renrui Zhang, Yufei Liu, Kai Wang, Wanli Ouyang, Zhiwei Xiong, Peng Gao, Qibin Hou, et~al.
\newblock Tar3d: Creating high-quality 3d assets via next-part prediction.
\newblock \emph{arXiv preprint arXiv:2412.16919}, 2024{\natexlab{b}}.

\bibitem[Zhang et~al.(2024{\natexlab{c}})Zhang, Yin, Chen, Lin, Li, Hou, and Cheng]{zhang2024temo}
Xuying Zhang, Bo-Wen Yin, Yuming Chen, Zheng Lin, Yunheng Li, Qibin Hou, and Ming-Ming Cheng.
\newblock Temo: Towards text-driven 3d stylization for multi-object meshes.
\newblock In \emph{Proceedings of the ieee/cvf conference on computer vision and pattern recognition}, pages 19531--19540, 2024{\natexlab{c}}.

\bibitem[Zhao et~al.(2023)Zhao, Liu, Chen, Zeng, Wang, Cheng, Fu, Chen, Yu, and Gao]{zhao2023michelangelo}
Zibo Zhao, Wen Liu, Xin Chen, Xianfang Zeng, Rui Wang, Pei Cheng, Bin Fu, Tao Chen, Gang Yu, and Shenghua Gao.
\newblock Michelangelo: Conditional 3d shape generation based on shape-image-text aligned latent representation.
\newblock \emph{Advances in neural information processing systems}, 36:\penalty0 73969--73982, 2023.

\bibitem[Zheng and Vedaldi(2024)]{zheng2024free3d}
Chuanxia Zheng and Andrea Vedaldi.
\newblock Free3d: Consistent novel view synthesis without 3d representation.
\newblock In \emph{Proceedings of the IEEE/CVF Conference on Computer Vision and Pattern Recognition}, pages 9720--9731, 2024.

\bibitem[Zhou et~al.(2024)Zhou, Zhou, Cheng, Feng, and Hou]{zhou2024storydiffusion}
Yupeng Zhou, Daquan Zhou, Ming-Ming Cheng, Jiashi Feng, and Qibin Hou.
\newblock Storydiffusion: Consistent self-attention for long-range image and video generation.
\newblock \emph{arXiv preprint arXiv:2405.01434}, 2024.

\bibitem[Zhu et~al.(2024)Zhu, Wang, and Yang]{zhu_2024_gsror}
Zuo-Liang Zhu, Beibei Wang, and Jian Yang.
\newblock G{S}-{ROR}: 3{D} {G}aussian splatting for reflective object relighting via sdf priors.
\newblock \emph{arXiv preprint arXiv:2406.18544}, 2024.

\bibitem[Zhu et~al.(2025)Zhu, Yang, and Wang]{zhu_2025_dsdf}
Zuo-Liang Zhu, Jian Yang, and Beibei Wang.
\newblock Gaussian splatting with discretized sdf for relightable assets.
\newblock In \emph{Proceedings of IEEE International Conference on Computer Vision (ICCV)}, 2025.

\bibitem[Zuo et~al.(2024)Zuo, Gu, Qiu, Dong, Zhao, Yuan, Peng, Zhu, Dong, Bo, and Huang]{zuo2024videomv}
Qi Zuo, Xiaodong Gu, Lingteng Qiu, Yuan Dong, Zhengyi Zhao, Weihao Yuan, Rui Peng, Siyu Zhu, Zilong Dong, Liefeng Bo, and Qixing Huang.
\newblock Videomv: Consistent multi-view generation based on large video generative model, 2024.

\end{thebibliography}
}

\end{document}